\documentclass[letterpaper, 10 pt, conference]{IEEEtran}
\IEEEoverridecommandlockouts
\usepackage{cite}
\usepackage{amsmath,amssymb,amsfonts}
\usepackage{algorithmic}
\usepackage{graphicx}
\usepackage{textcomp}
\usepackage{xcolor}
\def\BibTeX{{\rm B\kern-.05em{\sc i\kern-.025em b}\kern-.08em
    T\kern-.1667em\lower.7ex\hbox{E}\kern-.125emX}}
    
\usepackage{hyperref}
\usepackage{dsfont}
\usepackage[inline]{enumitem}
\usepackage{multirow}
\usepackage{makecell}
\usepackage[caption=false]{subfig}

\usepackage{svg}

\usepackage{pifont}

\hypersetup{
  colorlinks   = true,    
  urlcolor     = blue,    
  linkcolor    = blue,    
  citecolor    = red      
}

\begin{document}

\title{Saving the Limping: Fault-tolerant Quadruped Locomotion via Reinforcement Learning
}

\author{Dikai Liu$^{1,2}$, Tianwei Zhang$^{2}$, Jianxiong Yin$^{1}$ and Simon See$^{1,3}$%
\thanks{$^{1}$ NVIDIA AI Technology Centre (NVAITC); e-mail: {\tt\small \{dikail,jianxiongy,ssee\}@nvidia.com}}
\thanks{$^{2}$ School of Computer Science and Engineering, Nanyang Technological University, Singapore; e-mail: {\tt\small dikai001@e.ntu.edu.sg, tianwei.zhang@ntu.edu.sg}}
\thanks{$^{3}$ also with Coventry University and Mahindra University}
}

\maketitle

\begin{abstract}
Modern quadrupeds are skillful in traversing or even sprinting on uneven terrains in a remote uncontrolled environment. However, survival in the wild requires not only maneuverability, but also the ability to handle potential critical hardware failures. How to grant such ability to quadrupeds is rarely investigated. In this paper, we propose a novel methodology to train and test hardware fault-tolerant controllers for quadruped locomotion, both in the simulation and physical world. We adopt the teacher-student reinforcement learning framework to train the controller with close-to-reality joint-locking failure in the simulation, which can be zero-shot transferred to the physical robot without any fine-tuning. Extensive experiments show that our fault-tolerant controller can efficiently lead a quadruped stably when it faces joint failures during locomotion.

\end{abstract}


\section{Introduction}

Benefiting from the rapid advances in hardware and control algorithms, quadrupedal robots are becoming more intelligent in solving various tasks with good performance. They demonstrate high flexibility and versatility in complex contexts, and are expected to tackle many critical real-world missions, such as search \& rescue \cite{bellicoso2018advances}, patrol \cite{chen2021autonomous} and delivery \cite{hooks2020alphred}. 

Quadrupeds are normally deployed in remote uncontrolled environments \cite{miki2022learning}, where accidents could happen at any time to cause potential critical hardware failures to the physical device, e.g., joint locking, free swinging, broken brackets. These failures could bring significant harm to the robots and humans, increase the down time, and shorten the service life of the robots. Therefore, it is important for the onboard controller to be robust against the hardware failures and bring the quadrupeds back home safely. The nature of quadruped instability makes it more susceptible to failures compared to other robotic platforms \cite{chen2022fault} and fault tolerance is a crucial aspect in the design of quadruped controllers. 

Unfortunately, existing commercial quadrupeds employ limited hardware failure detection (e.g., motor overheating, sensor signal loss) or protection functions (e.g., shutting down the system), which are not sufficient to operate the system safely and effectively in dynamic and unpredictable outdoor environments. In the research community, previous studies have introduced solutions to achieve various locomotion tasks under \textit{normal conditions}, such as traveling through rough terrains \cite{lee2020learning, kumar2021rma, miki2022learning}, jumping \& falling recovery \cite{park2021jumping}, running at high speeds \cite{margolisyang2022rapid} and dexterous manipulation \cite{shi2021circus}. How to cope with hardware failures at runtime without disruption to the system is still an unsolved problem. 

It is challenging to grant the quadrupeds the capability of handling hardware failures automatically. Traditional control theory methods, such as the model predictive control (MPC) and whole body control (WBC) frameworks, require manual tuning of model parameters with in-depth domain knowledge \cite{di2018dynamic,gehring2016practice}. Using such a predetermined motion and trajectory planner is significantly restricted in the real world, especially when facing unknown environmental conditions and failures: 
even different situations of the same failure type (e.g., joint locking) need to be modeled separately \cite{cui2022fault}, making them more difficult to generalize and scale to different failure types and hardware configurations.


A more promising strategy is to apply reinforcement learning (RL) algorithms to train the policy in a simulator, and then transfer it to the physical world \cite{kumar2021rma, margolisyang2022rapid, lee2020learning}. This can remarkably relax the requirement of domain knowledge. To improve the model performance in the real world, a number of simulators like Isaac Gym \cite{makoviychuk2021isaac}, have been developed with photo-realistic rendering and physical-accurate modeling. Meanwhile, many techniques are proposed to reduce the sim-to-real gap \cite{tobin2017domain, peng2018sim, hwangbo2019learning}. However, such gap still exists especially in the context of hardware failures for two reasons. First, it is impossible to simulate every possible environment state where hardware failures could occur, even with the domain randomization technique \cite{tobin2017domain}. The consideration of too many environment states can significantly slow down the training process, or even cause convergence failures. Second, the physical robot model used in the simulator is usually simplified, and cannot reflect the real robot conditions (e.g., with hardware faults). Recent works proposed several RL-based controllers to achieve fault-tolerance \cite{anne2021meta,okamoto2021reinforcement}, which are only tested in the simulation environment. Due to the huge sim-to-real gap, it is unclear how these methods will perform in the physical world.

\begin{figure}[t]
    \centering
    \includegraphics[width=.8\columnwidth]{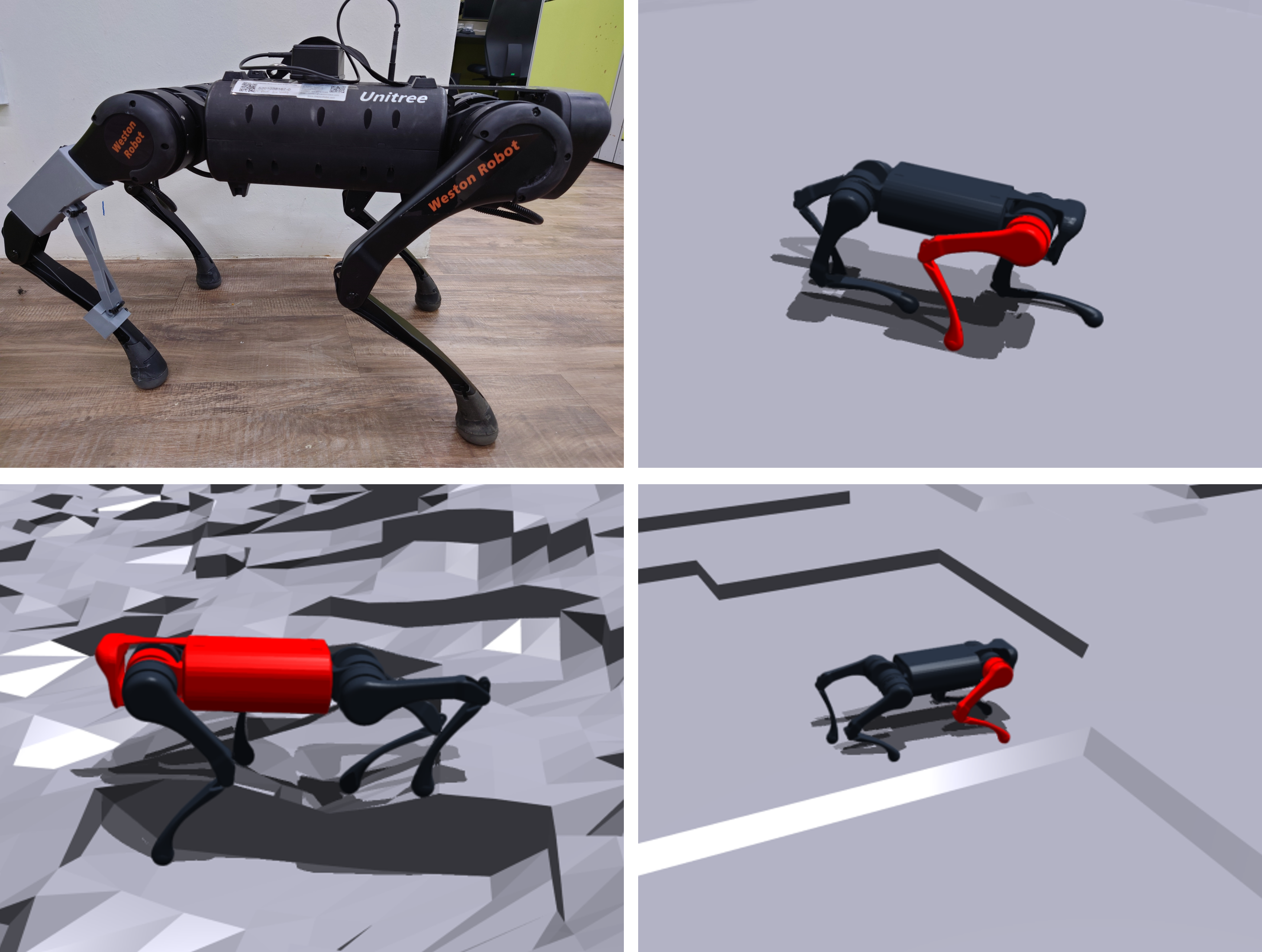}
    \caption{Physical robot and its simulated counterpart. Unitree A1 is equipped with our joint locking mechanism. Its official URDF model is used in the Isaac Gym simulator \cite{makoviychuk2021isaac} with body links of the locked joint showing in red.}
    \vspace{-10pt}
    \label{fig:robots}
\end{figure}


Motivated by these limitations, we design a novel framework to achieve robust fault-tolerance quadruped locomotion in the physical world. We make the following contributions. (1) We design a simulation strategy to realistically simulate joint locking failures, and a locking mechanism for real-world testing. The randomized failure in simulation helps to train a generalized agent that can handle various joint locking scenarios in the real world, instead of certain pre-definded cases. (2) We adopt the teacher-student reinforcement learning paradigm to achieve jointly single-phase training and zero-shot transfer. The student model can efficiently extract information from the onboard sensors. When deployed in a physical quadruped, the policy can provide real-time locomotion control against possible hardware failures in uncontrolled environments. We conduct extensive experiments in both simulation and a physical Unitree AI robot (Fig. \ref{fig:robots}). Evaluations show that our method can significantly improve the robustness and hardware fault tolerance of RL-based control policies. 

\section{Related Work}

\subsection{RL-based Quadruped Locomotion}
Prior works introduced human-designed quadruped MPC controllers for real-world applications \cite{di2018dynamic, neunert2018whole}. Recent research in reinforcement learning provides an alternative direction to the design of robotic controllers with less prior knowledge. For instance, Rudin et al. \cite{rudin2022learning} adopted multiple types of rough terrains to improve the robustness of the quadruped agent. Kumar et al. \cite{kumar2021rma} developed an adaption framework for locomotion. This framework was further extended in \cite{margolisyang2022rapid} to set a new world record for quadruped high-speed running. More advanced skills such as wheel-based locomotion \cite{vollenweider2022advanced} and limb control \cite{shi2021circus} reveal the unlimited potential of RL algorithms for quadruped locomotion and control.

\subsection{Sim-to-Real Gap in Robotic Control}
Sim-to-real gap \cite{koos2010crossing, boeing2012leveraging} has become a major obstacle for deploying RL-based algorithms. To reduce such gap, the most direct way is to conduct more physical-accurate and photo-realistic simulation \cite{makoviychuk2021isaac} or use real data to tune the virtual model \cite{tan2018sim}, which are not always available. 
Another common approach, known as domain randomization (DR) \cite{tan2018sim, tobin2017domain}, is to randomize parameters during simulation. Loquercio et al. \cite{loquercio2019deep} applied DR on the environmental texture to train a racing drone. Andrychowicz et al. \cite{andrychowicz2020learning} trained dexterous in-hand manipulation with random physical properties and object status.
Chebotar et al.\cite{chebotar2019closing} used real-world information to tune the randomization of simulations. In quadruped locomotion, Tan et al. \cite{tan2018sim} applied random noise on sensor data, hardware specifications, and environmental factors, which is widely adopted by subsequent works. Randomly generated terrains \cite{kumar2021rma, rudin2022learning, lee2020learning} are now commonly used to improve the robustness of the policy against different environments. However, over randomization can cost optimality and introduces a trade-off between policy performance and training speed, leading to an over-conservative policy or even training collapse \cite{luo2017robust}.

\subsection{Quadruped Fault Tolerance}
Despite the wide application of quadrupedal robots, there are only a few works studying the fault-tolerant control \cite{zhong2019analysis}. They can be classified into the following two categories, and each one suffers from some limitations. 

The first strategy is to use traditional control theory methods, which has the generalization issue. Specifically, some papers considered the single joint locking failure, and adopted the classic control theory to develop fault-tolerant gaits and inverse kinematics solutions \cite{yang2006kinematic, pana2008fault, gor2018fault}. Recently, Cui et al. \cite{cui2022fault} proposed a whole-body control (WBC) method to optimize the posture to handle joint lock failures. However, most of these methods are only designed for certain specific quadrupedal robots, and require intensive manual analysis and modeling for different scenarios. To design an intelligent fault-tolerant controller, Koos et al. \cite{koos2013fast} designed an algorithm for the hexapod robot to discover compensatory behaviors in unanticipated situations and search for efficient behaviors. However, it still needs to gather real-world data in the failure situation and requires a long time to discover new behaviors, which is not suitable to be deployed on modern quadrupedal robots for critical missions. 

The second strategy is to use RL for fault-tolerance robotic control, which suffers from the sim-to-real gap. Okamoto et al. \cite{okamoto2021reinforcement} introduced a fault-tolerant RL algorithm. However, this solution is only tested in the Ant-v2 environment in the simulator and its transferability to the physical world with a real quadrupedal robot is unknown. Anne et al. \cite{anne2021meta} used meta RL and included the locked motor and leg amputation as uncertainties. Similarly, it is only tested in the simulator with SpotMicro, and struggles to generalize to unseen situations such as joint lock on different legs.

\subsection{Teacher-Student Training}
Online system identification predicts the underlying status, usually from a history of past states and actions in robotics \cite{yu2017preparing}. The teacher-student framework is a common approach, where the teacher model uses the privilege information for the student model to infer \cite{kumar2021rma,lee2020learning}. The privilege information can be a combination of ground truth states such as a terrain map \cite{margolis2021learning, miki2022learning,lee2020learning}, and randomized domain parameters \cite{kumar2021rma, margolisyang2022rapid}. The student model learns to imitate the teacher model from perceivable noisy sensor input like IMU, and joint encoders \cite{kumar2021rma, margolisyang2022rapid} or with point cloud \cite{margolis2021learning,miki2022learning}.

\begin{figure*}[t]
\centerline{\includegraphics[width=0.8\textwidth]{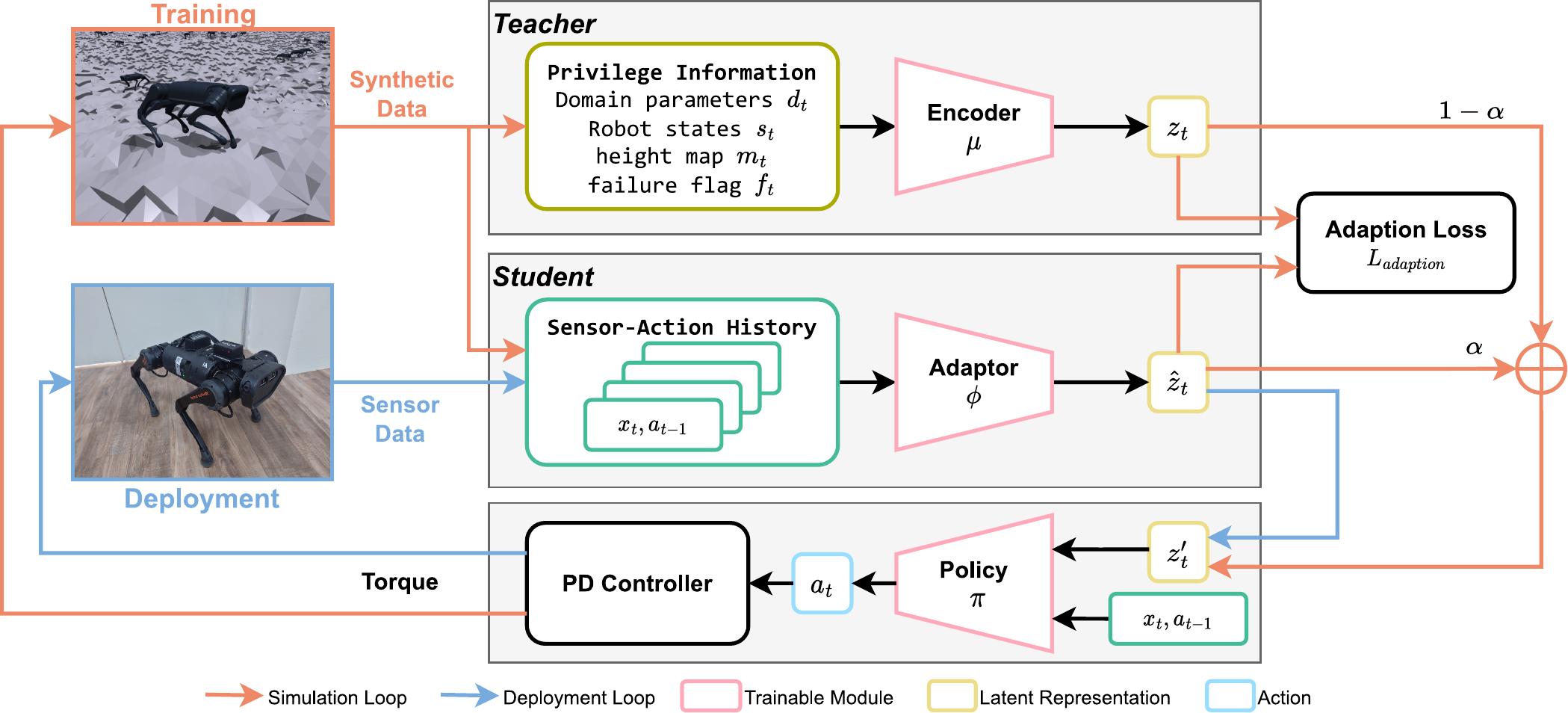}}
\caption{Overview of our methodology. We adopt the reinforcement learning architecture with the teacher-student framework from \cite{kumar2021rma,margolisyang2022rapid} to train the policy. The architecture consists of a teacher network $\mu$, a student network $\phi$, and a policy network $\pi$. During training, synthetic data from the simulator are used to compute the latent representation $z_t$ and $\hat{z}_t$ of the teacher and the student, respectively. By fusing the latent information, we train all three networks jointly for fast convergence in the early stage and then an optimized student policy in the end. The policy and student model will be directly deployed on the physical robot without any further offline training or fine-tuning. During deployment, policy network takes only $\hat{z}_t$ from student network as the latent representation.}
\vspace{-10pt}
\label{fig:framework}
\end{figure*}

\section{Methodology}

Our goal is to train a control policy $\pi$ to guide the stable locomotion of the quadruped even when it faces critical hardware failures (e.g., joint locking). This policy takes as input the latent representation $\hat{z}_t$ encoded from the perceivable sensor data and outputs the desired joint position. It is designed to be capable of zero-shot deployment in the real world. Fig.~\ref{fig:framework} presents the overview of our methodology. 

\subsection{Joint Locking in Quadruped Locomotion} \label{section:joint_failure}
Quadrupeds deployed in remote uncontrolled environments face the challenges of unpredictable joint failure, which can immobilize or even damage the robot \cite{cui2022fault}. Joint locking and free-swinging are the most common faulty situations. A locked joint cannot be controlled freely and has only a limited range of motion. However, the actuator can still support the body as torques are still applied. A free swing joint cannot be controlled and no actions are made. It moves easily by an external force, and thus cannot support the body. In this paper, we mainly focus on the single joint locking failure, which is also the target of recent related works \cite{cui2022fault, okamoto2021reinforcement, anne2021meta}.

\noindent\textbf{Joint Failure in Simulation.} To safely develop a fault-tolerant controller, we use Isaac Gym \cite{makoviychuk2021isaac} to simulate failure situations in the locomotion task with domain randomization. We first create a vanilla environment \texttt{BaseEnv}, where no failure occurs. Then for each virtual agent, we randomly sample the failure time $T_f \sim \mathcal{U}(T^f_{min}, T^f_{max})$, the failure joint $J_f \sim \mathcal{U} \{1, \dots, 12 \}$ and the failure tolerance $\theta_{tol} \sim \mathcal{N}(0, {\theta_{max}}^2) $. The failure status is tracked by a failure flag $f_t \in [0, 12]$. 

Initially and after every reset, $f_t$ is cleared as 0 to indicate a normal state. In the episode, when the agent progresses to $T_f$, the failure occurs, and $f_t$ is updated to reflect the joint failure $f_t = J_f$. The current position of the selected joint $J_t$ is used as the central failure angle $\bar{\theta} = q_{J_t}$. We model joint locking failure by restricting the joint movement with a limited range $\theta_{allowed}$, controlled by the central position $\bar{\theta}$ and symmetric tolerance $\theta_{tol}$, which are used to directly overwrite the joint's limit with Isaac Gym's API.

We refer to the failure environment as \texttt{FailureEnv}. Unlike previous methods \cite{okamoto2021reinforcement, anne2021meta}, where joint failures are predefined and fixed, we use domain randomization to generate versatile and unpredictable situations. Since joint locking directly affects joint control, and the robot status and surroundings at the failure moment can greatly alter the result, online randomization can help to train a robust and generalized policy against various joint locking accidents.

\noindent\textbf{Joint Failure in the Real World.}\label{section:physical_failure} Quadrupedal robots are complex machines, and modifying the hardware can be dangerous without the support from the manufacturer, especially when joint failure is intentionally added to the system. To safely evaluate the fault-tolerant controller on the physical robot, we use both hardware and software methods to simulate the joint locking situation in the real world.

\begin{figure}[t]
    \centering
    \subfloat[]{%
        \includegraphics[width=0.55\columnwidth]{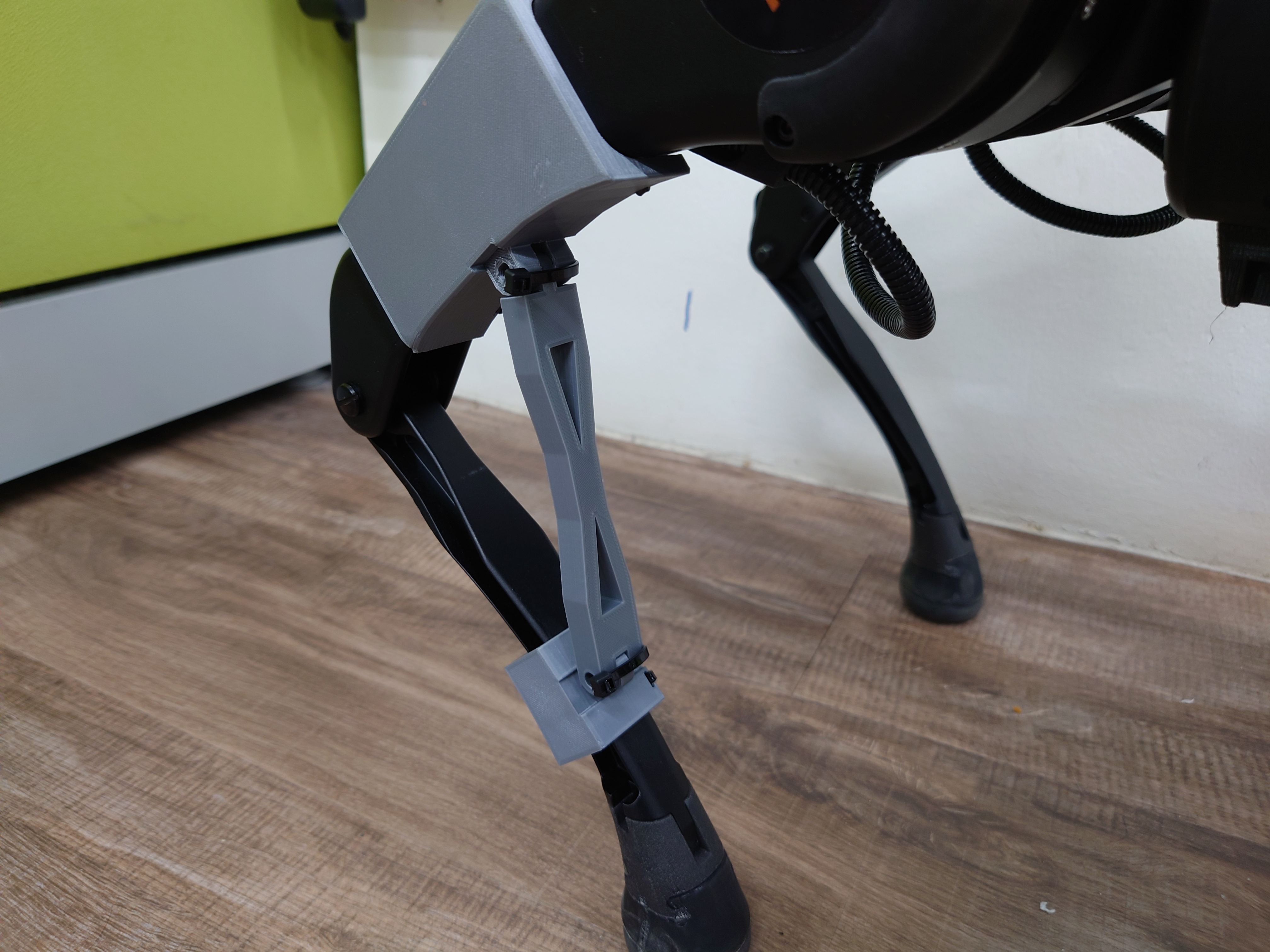}
        \label{fig:locking_assemble}
    }
    \hfill
    \subfloat[]{%
        \includegraphics[width=0.4\columnwidth]{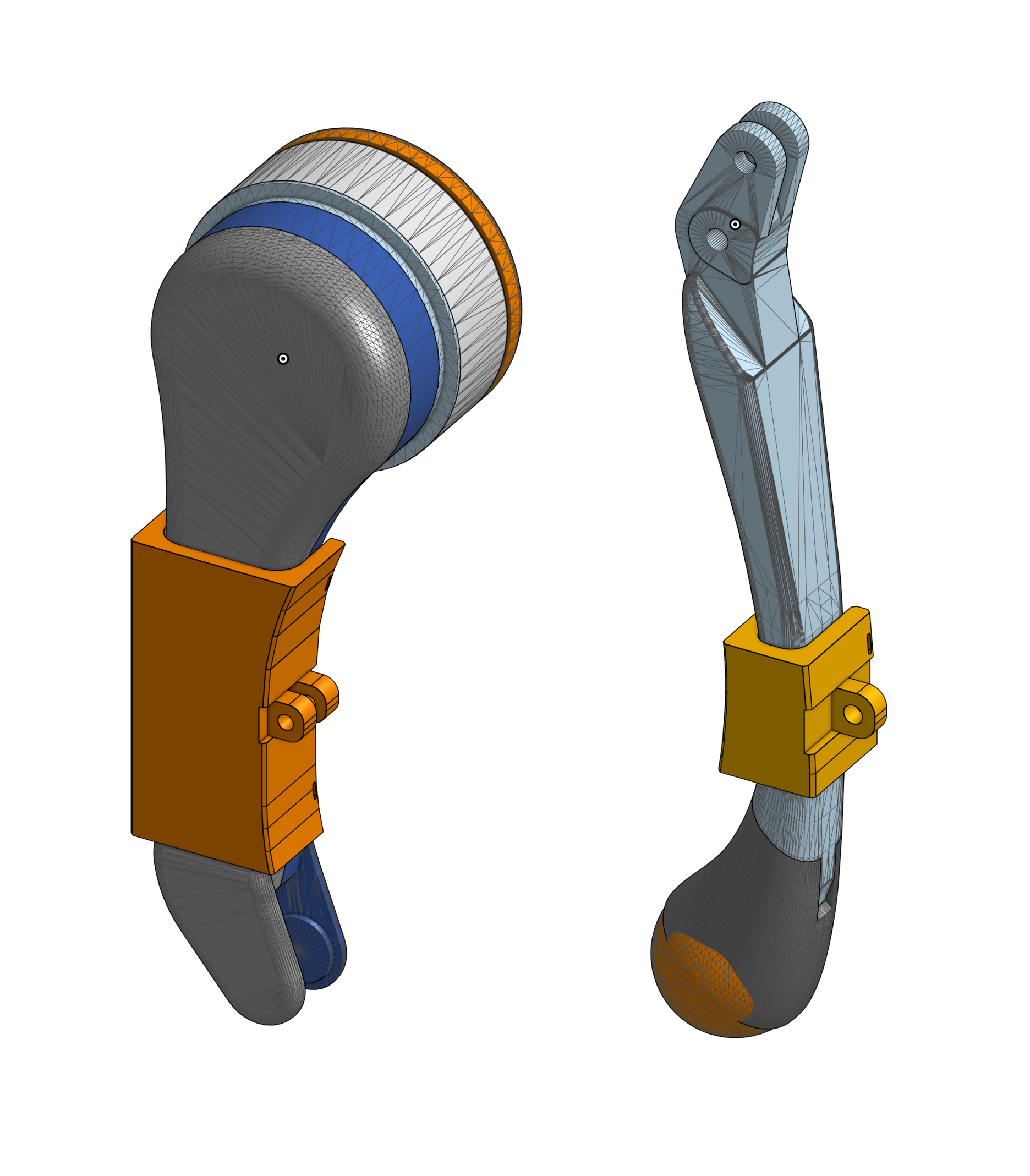}
        \label{fig:locking_3d}
    }
    \caption{The 3D-printed joint locking mechanism assembled in the physical device \protect\subref{fig:locking_assemble}, containing two mounts for thigh link and calf link \protect\subref{fig:locking_3d} for rod connection to form a locking situation.}
    \label{fig:joint_locking}
    \vspace{-10pt}
\end{figure}

For \textit{hardlock}, we design and 3D print an external locking mechanism (Fig.~\ref{fig:joint_locking}) to directly limit the motion of the joint. Although such locking is closer to the real situation, it cannot be easily used on every joint due to the hardware design of the quadruped. For most quadruped, such as Unitree A1, only the calf joint of each leg is exposed in the open space and the mechanism can be attached before the experiment, which could limit the diversity of test cases.

To address this limitation, we also use \textit{softlock} to simulate joint locking. Similar to the simulator, we track the joint position after failure occurs. The desired joint position from the controller is clipped in the range of $\theta_{allowed}$ before being passed to the onboard PD controller for torque output. With \textit{softlock}, joint failure can occur at any time and at any joint during the experiment. However, with a modified command, the real-world simulation can be different from the actual situation. Both methods are tested and evaluated in Sec.~\ref{section:physical_validataion}



\subsection{Reinforcement Learning Architecture}
To train a fault-tolerant quadruped controller, we adopt RL, which takes data from the common onboard sensor as observations, and outputs the optimal actions as joint position. 

\noindent\textbf{Observation.}
We collect data from the equipped low-level sensor to provide observations. At any time $t$, joint encoders, IMU and foot encoders provide noisy sensor data $x_t \in \mathbb{R}^{30}$. We further add the previous actions $a_{t-1} \in \mathbb{R}^{12}$ to form the observation $o_t = [x_t, a_{t-1}] \in \mathbb{R}^{42}$. Following recent works \cite{kumar2021rma, margolisyang2022rapid, lee2020learning}, we use a historical observations of length $H$ to capture the temporal information.

\noindent\textbf{Action.} The control policy $\pi$ predicts the target joint position $\hat{q} = a_t \in \mathbb{R}^{12}$, which is consequently processed by a PD controller for the desired torque $\tau$ of each actuator.
\begin{equation}
    \tau = K_p (\hat{q}-q) + K_d (\hat{\dot q}- \dot q) \nonumber
\end{equation}
where $K_p$ and $K_d$ are the stiffness and damping gain controlled by DR. The target joint velocity $\hat{\dot q}$ is set to 0.

\noindent\textbf{Reward Function.} Closely following \cite{kumar2021rma, margolisyang2022rapid}, the reward functions encourage the agent to move forward stably and smoothly with a target speed of 0.5 m/s. We mainly penalize movement in other axes, large acceleration, power consumption, and collision. 


\noindent\textbf{Domain Randomization.} Besides the failure simulation described in Sec.~\ref{section:joint_failure}, we also randomize ground friction, PD controller settings, payload, and motor strength to add robustness for various situations.

\subsection{Joint Teacher-Student Framework} \label{section:joint_training}
Our goal is to obtain a zero-shot policy, which is trained completely in the simulator and transferred to the real world without fine-tuning. For RL-based policies, the privileged underlying states of the robot and environment can produce better performance in a smaller number of training iterations \cite{tassa2018deepmind, laskin2020curl}. We adopt the teacher-student learning paradigm \cite{lee2020learning, kumar2021rma, margolisyang2022rapid} to achieve this goal. It enables implicit identification of the hidden dynamics of the environment and robots $e_t$ for different behaviors. It also learns dynamics from perceivable data, making the policy deployable in the real world.

Specifically, a teacher model $\mu$ is introduced to encode the environmental factor $e_t$ into the latent space representation $z_t$ with length $D$:
\begin{equation}
    z_t = \mu(e_t) \in \mathbb{R}^{D} \nonumber
\end{equation}
To better capture the dynamics, $e_t$ contains necessary underlying ground-truth synthetic data that are accessible from the simulator including the DR parameters $d_t$, clean robot states $s_t = [v_t, \omega_t]$ and height map $m_t$ of the surrounding terrain.

A student model $\phi$ is also introduced to learn from historical observations to mimic the encoding from $\mu$ by performing system identification:
\begin{equation}
    \hat{z}_t = \phi(o_{t-H:t}) \in \mathbb{R}^{D} \nonumber
\end{equation}

To optimize $\phi$, previous works \cite{kumar2021rma,lee2020learning,margolisyang2022rapid} focus on imitating $\mu$'s behaviors only by using supervised learning inspired by DAgger \cite{ross2011reduction}. With the trajectory generated by trained $\mu$ or randomly initialized $\phi$ in the online or offline fashion, $\mathcal{L}_{adaption} = \| z_t - \hat{z}_t \| ^2$ is minimized and used with previously trained $\pi$. However, it is almost impossible to get an exact latent representation such that $\hat{z}_t = z_t$. The difference in the latent representation can cause unpredictable behaviours, and performance degradation. To address this issue and minimize uncertainty, we propose to fuse the output of $\mu$ and $\phi$ with an adaptive ratio $\alpha$ to jointly optimize the policy network with the student network:
\begin{gather}
    z'_t = \alpha \hat{z}_t + (1 - \alpha) z_t \nonumber \\
    a_t = \pi(z'_t, o_t) \nonumber
\end{gather}

We update $\alpha$ with the progression of policy optimization. In the early stage, we set $\alpha=0$ to train $\pi$ by leveraging the privilege information encoded by $\mu$. With the training going on, we gradually increase $\alpha$, until only $\hat{z}_t$ is used for policy making in the late stage. Thus, even if we cannot get a perfect replica of $z_t$, $\pi$ can still have the opportunity to learn to adapt to such a difference in a single phase to maximize the reward.

To train the adaption jointly, we append the PPO \cite{schulman2017proximal} loss $\mathcal{L}_{RL}$  with $\mathcal{L}_{adaption}$, which is similar to \cite{wu2021self,radosavovic2023learning}:
\begin{equation}
    \mathcal{L} =  \mathcal{L}_{RL} + \beta \mathcal{L}_{adaption} \nonumber
\end{equation}
The adaptive ratio $\beta$ is negatively correlated with $\alpha$. As the proportion of the student output increases for policy making, we can focus more on the reward benefits rather than mimicking the teacher's behavior.

\section{Evaluation}

\subsection{Implementation and Experimental Setup}

\noindent\textbf{Module Implementation.} Both the teacher model $\mu$ and control policy $\phi$ are implemented in MLP with hidden layers of $[512, 256, 128]$ and $[256, 128]$, respectively, and ELU activation. $\mu$ outputs the latent representation with length $D=8$. Following \cite{kumar2021rma}, the student model adopts 1D CNNs to capture temporal information with a history length $H=50$ followed by a linear projection to the same latent space. All models are optimized jointly with PPO \cite{schulman2017proximal} as described in Sec.~\ref{section:joint_training}.

\noindent\textbf{Simulation.} Isaac Gym and its open source library IsaacGymEnvs \cite{makoviychuk2021isaac} are used to simulate massive parallel environments with rough terrains, including rough sloped terrain, smooth sloped terrain and discrete obstacles \cite{rudin2022learning}. We run the simulation on two NVIDIA A6000 GPUs, each handling 4096 environments at 200Hz, which can provide more than 0.1M FPS for simulation. The controller runs at 50Hz for command. 


\noindent\textbf{Hardware.} We adopt Unitree A1 as the test platform. It is a low-cost quadruped driven by 12 direct-drive actuators, equipped with IMU and foot-end force sensors used as observations. We use an external NVIDIA Jetson Xavier NX to replace the onboard Raspberry Pi high-level controller for GPU acceleration to process the exported JIT model.


\noindent\textbf{Agents and Baselines.}
To evaluate the impact of joint locking, we train different variants of the following \textbf{[T]}eacher networks:
\begin{itemize}[leftmargin=*]
    \item \textbf{BaseEnv[T]}: The teacher network trained in vanilla \texttt{BaseEnv} with privilege information.
    \item \textbf{FailureEnv[T]}: The teacher network trained in \texttt{FilureEnv} with privilege information as a baseline of fault-tolerant performance.
    \item \textbf{FailureEnv[T] w/o FF}: A variant of \textbf{FailureEnv[T]} where failure flag (FF) is removed from the privilege information. 
\end{itemize}
Each teacher model is trained from scratch with the same environment and PPO configuration.

To further evaluate the efficiency of knowledge transfer and performance in terms of reward, forward velocity, and survival time for fault-tolerant control, we train the following \textbf{[S]}tudent policies in \texttt{FailureEnv} with both proposed joint training (\textbf{[JT]}) and separate supervised (\textbf{[SS]}) from RMA \cite{kumar2021rma}:
\begin{itemize}[leftmargin=*]
    \item \textbf{\texttt{BaseEnv}[S][SS].} Student agent with supervised transfer but without fault-tolerance as a baseline for vanilla quadruped policy 
    \item \textbf{\texttt{BaseEnv}[S][JT].} Joint optimized variant for vanilla agent.
    \item \textbf{\texttt{FailureEnv}[S][SS].} Supervised transfer variant for the proposed method.
    \item \textbf{\texttt{FailureEnv}[S][JT].} Joint optimized transfer for the proposed fault-tolerant policy.
\end{itemize}
For fair comparisons, every student policy is trained with the same number of simulation steps. For \textbf{[SS]} policy, we equally divided the steps for the teacher and supervised training stages.

\begin{figure}[t]
    \centering
    \subfloat[Reward return for teacher policies]{%
        \includegraphics[width=0.9\columnwidth]{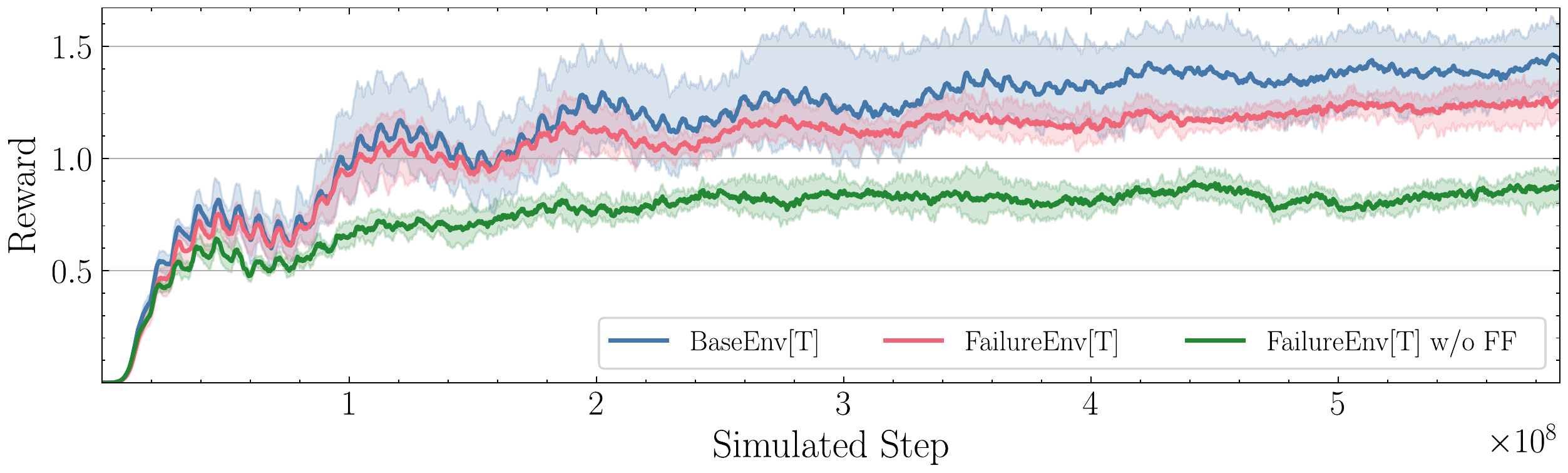}
        \label{fig:reward_impact}
    }
    \\
    \subfloat[Reward return for student policies]{%
        \includegraphics[width=0.9\columnwidth]{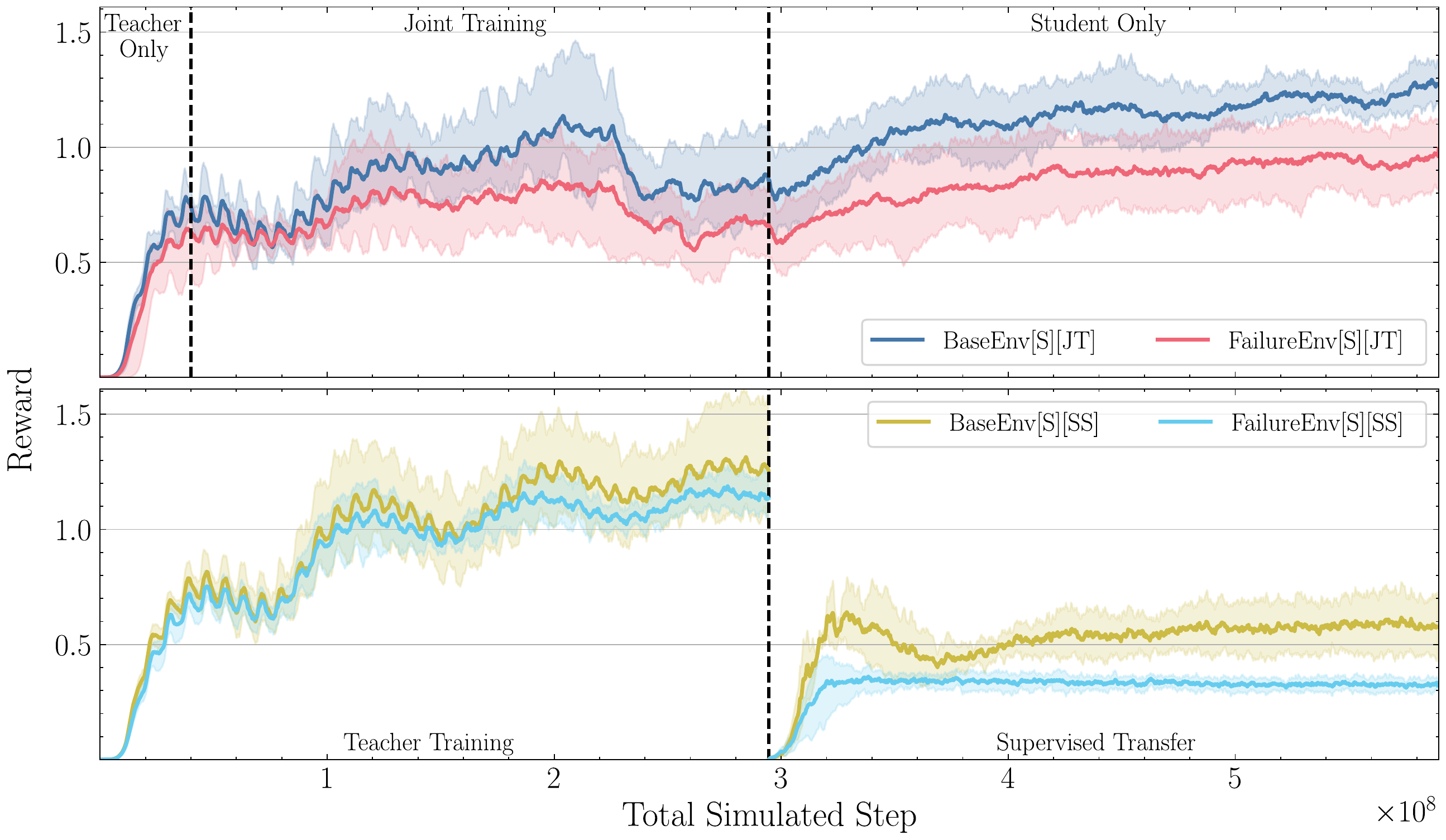}
        \label{fig:teacher_student_transfer}
    }
    \caption{Reward return in training different teacher and student policies.}
    \vspace{-10pt}
    \label{fig:training_reward}
    
\end{figure}

\subsection{Joint Locking Failure and Privilege information Impact} \label{section:locking_impact}

To understand the impact of failure and importance of joint failure information, we use \textbf{[T]} policies to eliminate the affect from knowledge transfer. We train each agent multiple times with different seeds using the privilege information and collect 600M simulated steps to ensure the best performance is achieved. Fig.~\ref{fig:reward_impact} shows the average reward for each agent.

When introducing joint locking failures into the virtual environment, we see a clear return drop from \texttt{BaseEnv}. This is expected since running in a degraded situation directly affects the speed and stability, which are the key parts of the reward function. However, with the proper privilege information like failure flag $f_t$, the teacher policy can still effectively learn the policy and minimize the impact of joint failure.

\subsection{Teacher-Student Transfer}
The teacher agent cannot be directly deployed on the physical robot due to the usage of privilege information, which is either unavailable or expensive to acquire in real world. So knowledge transfer is needed to train the student policies. Fig.~\ref{fig:teacher_student_transfer} shows the tracked reward during the joint training transfer and separate supervised transfer from RMA \cite{kumar2021rma}.

For supervised transfer, we unroll the checkpoint from the midpoint of previous teacher policy to generate trajectories and ground truth. There is a signification return drop under such a limited number of total simulation steps, especially for \texttt{FailureEnv}, indicating that pure supervised transfer may not be able to capture the unpredictable dynamics with joint failure efficiently. With more steps in both stages, the final performance will be better, but it requires much more time for simulation and training. The added unpredictability of joint locking also makes supervised learning transfer much worse, while the proposed joint training can still efficiently utilize the privilege information and transfer to the student. 

For joint training, the teacher network is trained solely in the first few epochs to generate a teacher policy that starts to drive the quadruped to walk. $\mathcal{L}_{adaption}$ and $\hat{z}_t$ then starts to fuse into the loss function and policy making until the midpoint of the overall prograss where student policy completely takes over and we focus on maximizing the reward again. Compared to supervised transfer, joint training can effectively transfer knowledge for both environments and achieve a higher reward return. This adds robustness and shows satisfactory performance for practical deployment. 

\subsection{Virtual Deployment}
\noindent\textbf{Overall Performance.} We deploy both \texttt{BaseEnv} and \texttt{FailureEnv} student policies into the same test environment where robots are spawned across different terrains and levels evenly with joint locking failure occurs randomly. Each virtual robot can run a maximum of 20 seconds after joint failure occurs. We track the forward velocity both before and after joint locking. We also measure the survival time of each agent on average, 25\% percentile (P25) and 50\% percentile (P50) so that we can see how each agent handles joint locking in the worst scenarios. The result averaged over 1500 instances per terrain is shown in Table~\ref{table:agent_performance}.

Before joint failure, both agents can drive the robot forward close to the target velocity of 0.5 m/s. After failure occurs, the velocity drops in both agents, but the fault-tolerant \texttt{FailureEnv} agent maintain the velocity slightly better. During deployment, the critical failure mostly kills the robot within seconds after joint locking, and the surviving instance can normally remain until the end, thus increasing the average survival time. Despite \texttt{BaseEnv} agent can still walk with a reasonable velocity after joint locking, it is more vulnerable to joint locking and fails within 5s for half of the instances. In contrast, the fault-tolerant \texttt{FailureEnv} agent can survival much longer with a locked joint. In smooth slope and rough slope terrains, even with joint failure, most of the robot can survive to the end of the journey. Robots in discrete obstacle terrain have significantly worse performance. Due to the small physical size of the A1 robot, it is too difficult for it to step up and down even under normal hardware conditions \cite{lai2023sim, agarwal2023legged}.

\begin{table}[t]
\caption{Agent performance with Joint Failure in Simulation.}
\vspace{-10pt}
\label{table:agent_performance}
\begin{center}
\resizebox{\linewidth}{!}{
\begin{tabular}{c|c|cc|ccc}
\Xhline{1.5pt}
\multirow{2}{*}{Agent}      & \multirow{2}{*}{Terrain} & \multicolumn{2}{c|}{\begin{tabular}[c]{@{}c@{}}Avg. Forward\\ Velocity (m/s)\end{tabular}} & \multicolumn{3}{c}{\begin{tabular}[c]{@{}c@{}}Survival Time (\%)\end{tabular}} \\ \cline{3-7} 
                            &                          & Before                                        & After                                       & Average                     & P25                     & P50                      \\ \Xhline{1.5pt}
\multirow{4}{*}{BaseEnv}    & Smooth Slope             & 0.56                                          & 0.45                                        & 51.4                        & 6.7                     & 36.5                     \\
                            & Rough Slope              & 0.55                                          & 0.41                                        & 44.4                        & 4.7                     & 20.7                     \\
                            & Discrete       & 0.54                                          & 0.41                                        & 40.8                        & 4.4                     & 17.45                    \\ \cline{2-7} 
                            & All                      & 0.55                                          & 0.42                                        & 44.7                        & 5.0                     & 21.35                    \\ \Xhline{1.5pt}
\multirow{4}{*}{FailureEnv} & Smooth Slope             & 0.59                                          & 0.52                                        & 68.3                        & 20.1                    & 100.0                    \\
                            & Rough Slope              & 0.57                                          & 0.47                                        & 59.1                        & 11.7                    & 81.0                     \\
                            & Discrete       & 0.55                                          & 0.44                                        & 45.8                        & 6.6                     & 31.0                     \\ \cline{2-7} 
                            & All                      & 0.57                                          & 0.47                                        & 56.5                        & 10.8                    & 59.0                     \\ \Xhline{1.5pt}
\end{tabular}}
\vspace{-15pt}
\end{center}
\end{table}

\begin{figure}[t]
    \centering
    \includegraphics[width=1\columnwidth]{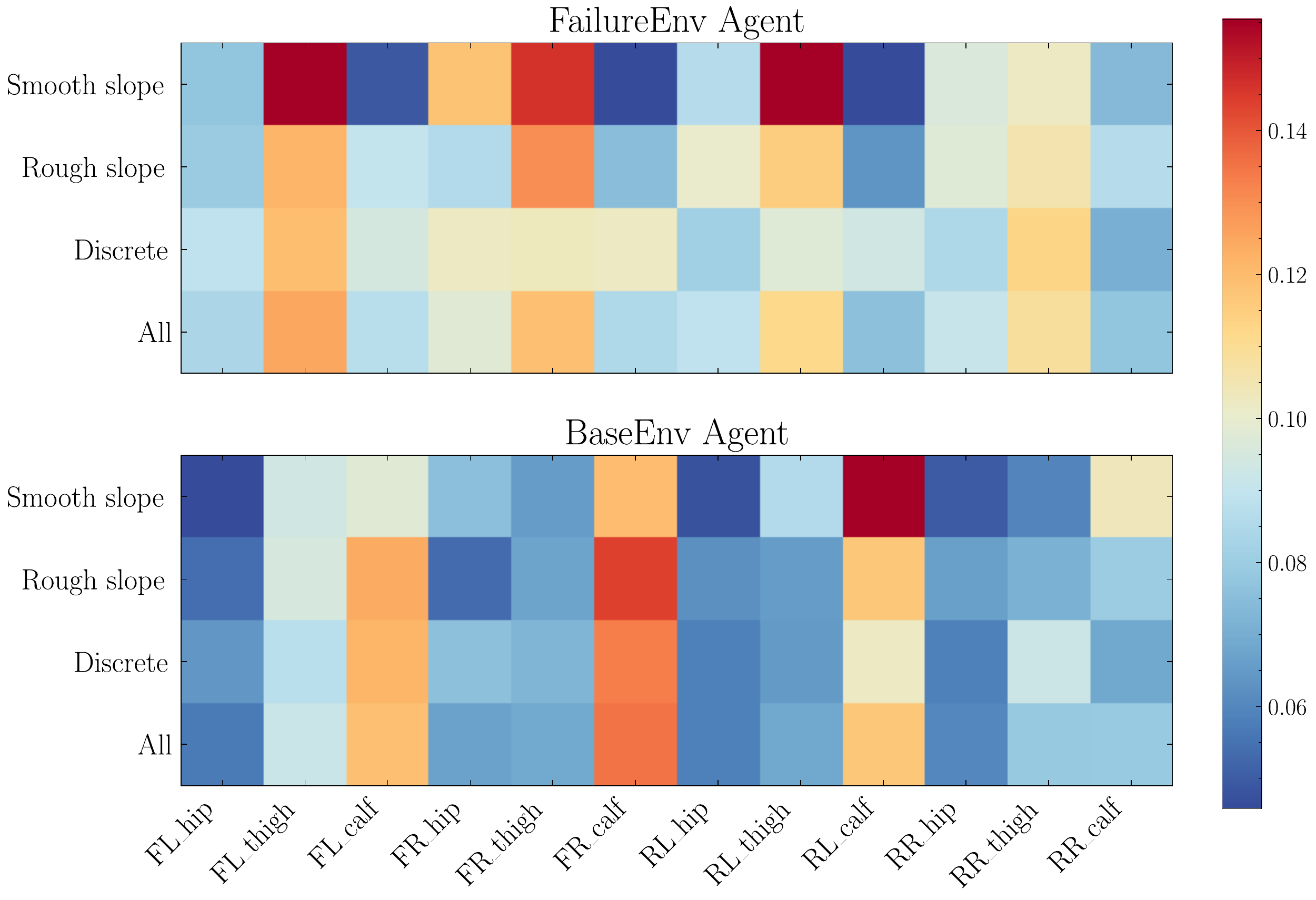}
    \caption{Joint distribution of failure joint in the worst cases of \texttt{FailureEnv} agent and \texttt{BaseEnv} agent}
    \label{fig:joint_dist}
\end{figure}

\noindent\textbf{Failure Case Study.}\label{section:virtual_failure} To further understand how joint locking affects control and leads to failure, we identify the most vulnerable instances of virtual deployment that fail within seconds for both agents. Fig.~\ref{fig:joint_dist} shows the distribution of the failure joint in the worst scenarios. The thigh and calf are the most vulnerable joint for \texttt{FailureEnv} and \texttt{BaseEnv} agents respectively. Both joints have larger movements compared to the hip joint, making them more sensitive to joint locking. The distribution shift reflects that while \texttt{FailureEnv} agent learns to overcome the locking of the calf joint, the thigh joint is still not fully handled. 

\begin{figure}[t]
    \centering
    \includegraphics[width=1.0\columnwidth]{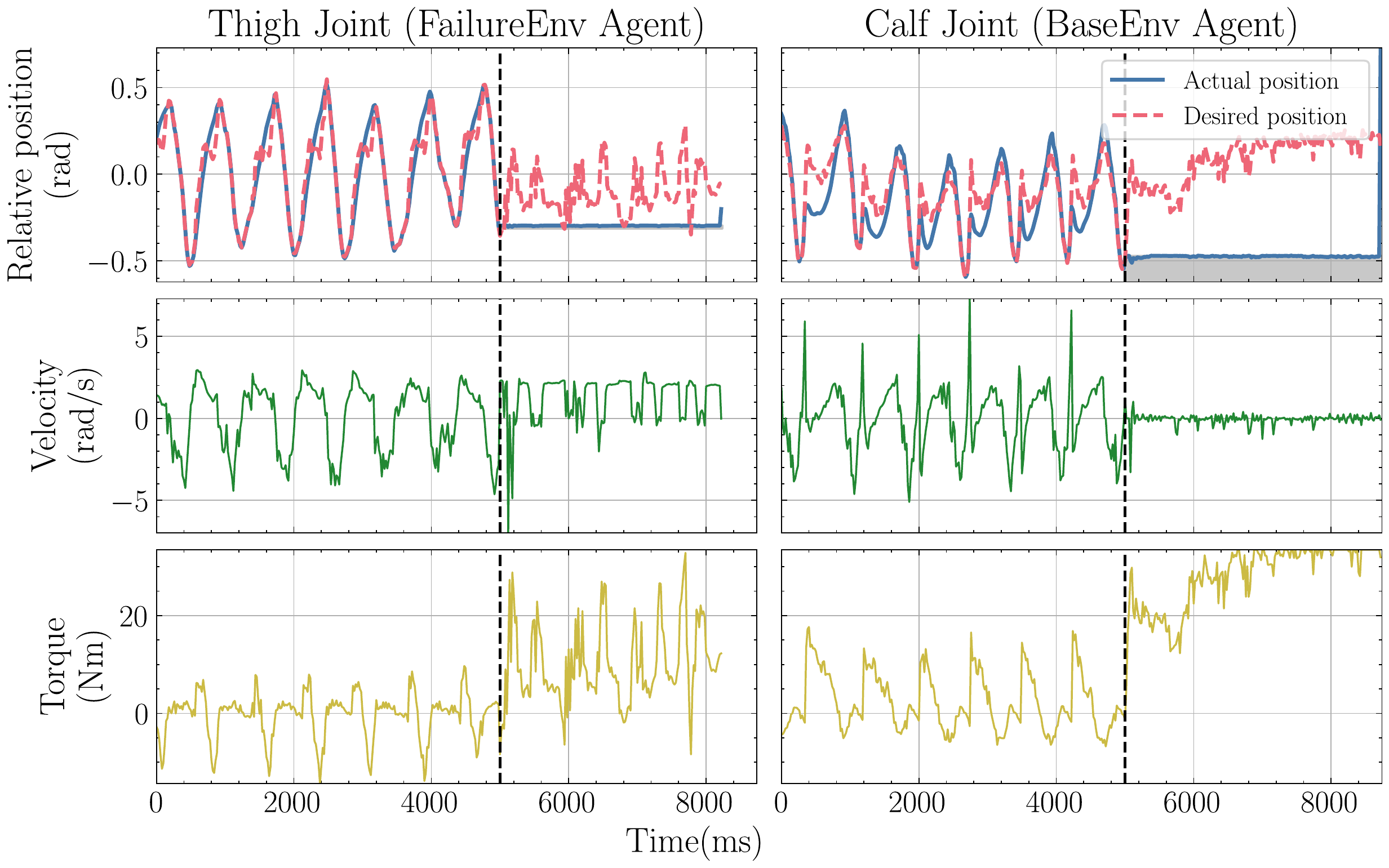}
    \caption{Motion of the most vulnerable joint for \texttt{FailreEnv} and \texttt{BaseEnv} agent identified in Fig.~\ref{fig:joint_dist}. We intercept the timeline from 5 seconds before joint locking to critical failure, where the agent needs to be reset. The grey box shows the limited motion range $\theta_{allowed}$. The joint position is relative to the default position for standing.}
    \vspace{-10pt}
    \label{fig:virtual_joint}
\end{figure}

We then track the joint status of these failure cases in Fig.~\ref{fig:virtual_joint}. Our proposed simulation strategy can effectively limit joint movement as desired in both position and velocity. With a locked joint, when the desired position is not in the range of $\theta_{allowed}$, we observe that a large torque is applied to the joint, which is another major factor of critical failure.

\subsection{Physical Validation} \label{section:physical_validataion}
\begin{figure*}[t]
    \centering
    \subfloat[Running with joint locking by the \texttt{FailureEnv} agent]{%
        \includegraphics[width=0.9\textwidth]{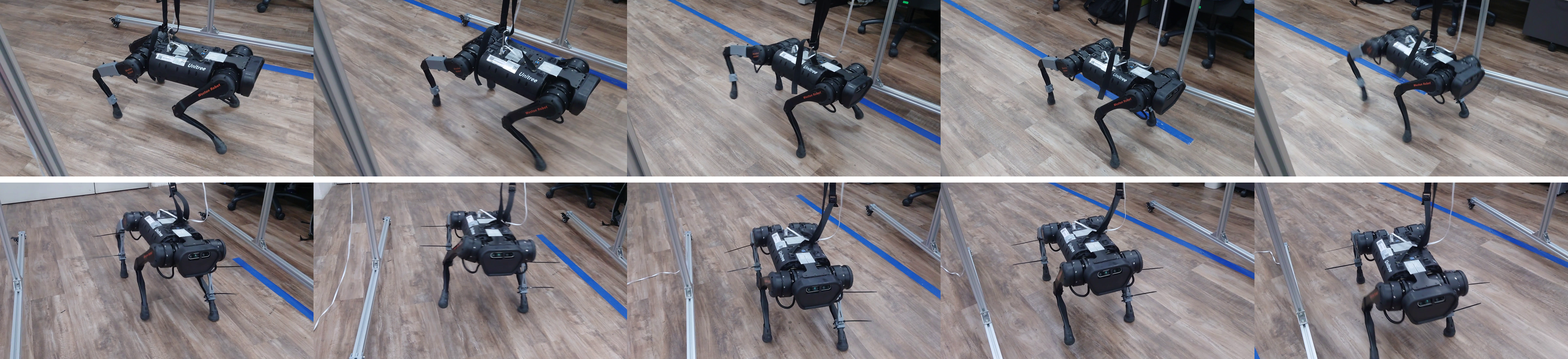}
        \label{fig:run_failure}
    }
    \\
    \subfloat[Running with joint locking by the \texttt{BaseEnv} agent]{%
        \includegraphics[width=0.48\textwidth]{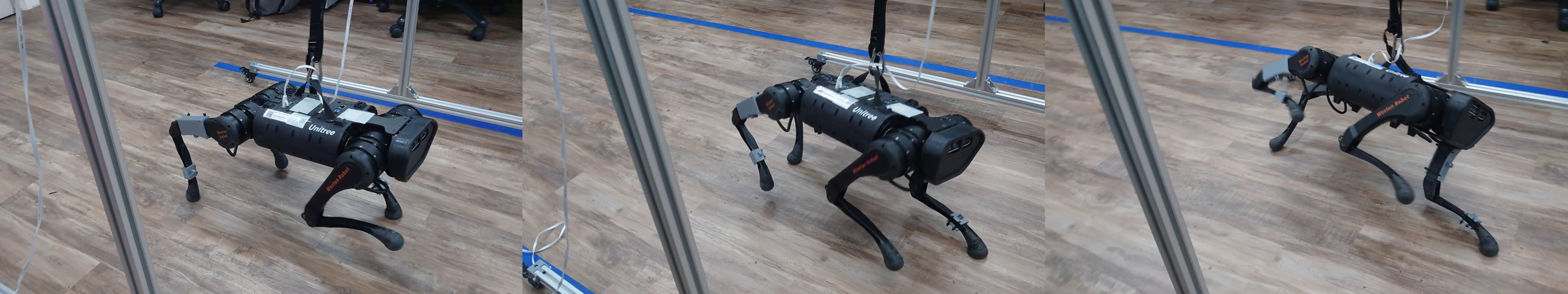}
        \label{fig:run_base}
    }
    \hfill
    \subfloat[Running with joint locking by built-in controller]{%
        \includegraphics[width=0.48\textwidth]{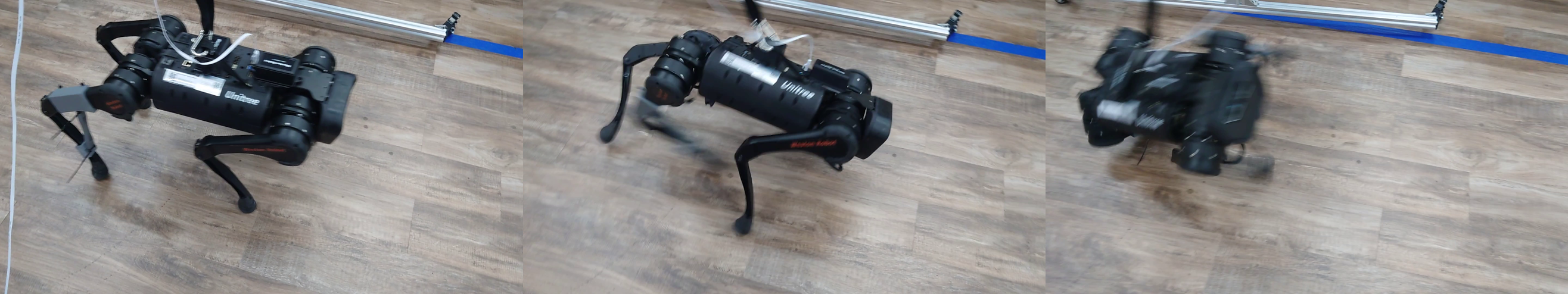}
        \label{fig:run_builtin}
    }
    \caption{Deployment snapshots on the physical robot run by \protect\subref{fig:run_failure} fault-tolerant \texttt{FailureEnv} agent, \protect\subref{fig:run_base} baseline \texttt{BaseEnv} agent and \protect\subref{fig:run_builtin} A1's built-in controller. The safety rope is only used to prevent hardware damage and does not affect running. Refer to the supplementary video for more information.}
    \label{fig:deployment}
\end{figure*}

We convert the trained model to JIT and deploy it directly on the physical Unitree A1 for zero-shot transfer without any fine-tuning. We compare the proposed fault-tolerant \texttt{FailureEnv} agent with the baseline model trained in \texttt{BaseEnv} and the built-in A1 controller. Following Sec.~\ref{section:physical_failure}, we use both \textit{hardlock} and \textit{softlock} during the deployment for comprehensive validation. Fig.~\ref{fig:deployment} shows the snapshots of the trials. 

\begin{figure*}[t]
\centering
\includegraphics[width=0.9\textwidth]{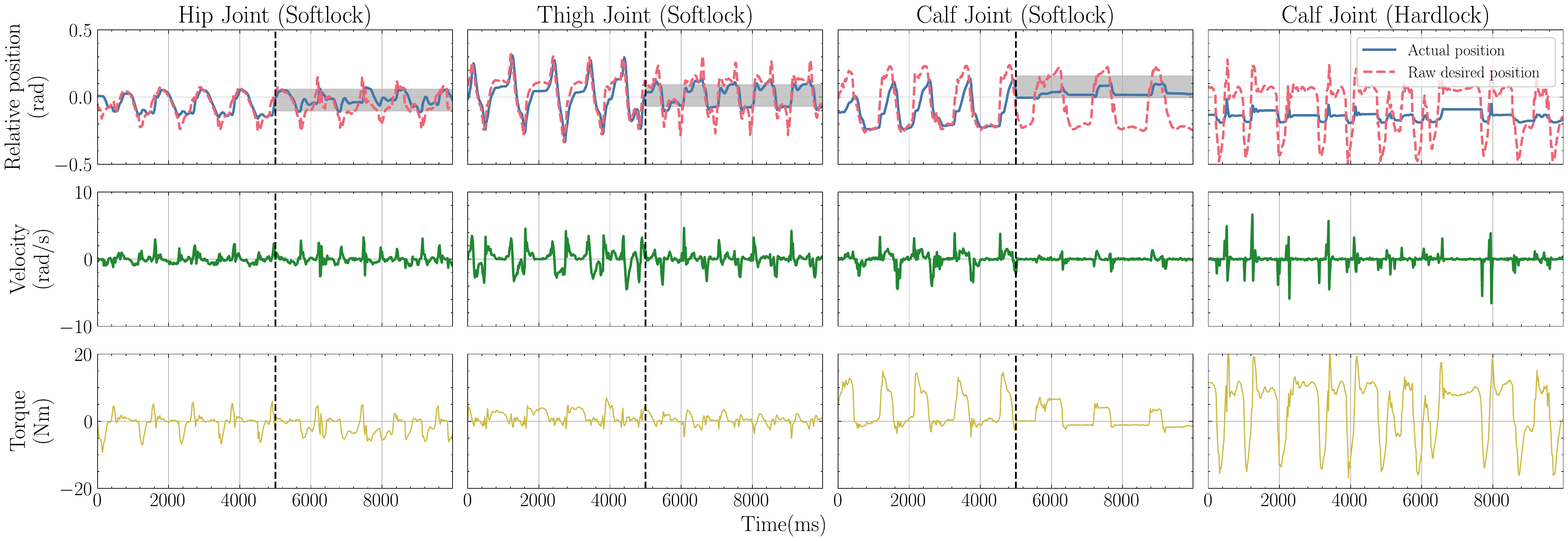}
\caption{Joint motion of the locked joint under both \textit{softlock} and \textit{hardlock} run by \texttt{FailureEnv} agent. For \textit{softlock}, we intercept 10 seconds around the joint locking timestamp, with the allowed movement range $\theta_{allowed}$ showing as a gray box. For \textit{hardlock}, the joint is locked at the beginning of the run. The joint position is relative to the default position for standing.}
\vspace{-10pt}
\label{fig:join_deployment}
\end{figure*}

Similar to virtual deployment, the joint motion is tracked during deployment. Fig.~\ref{fig:join_deployment} shows the motion of trails run by \texttt{FailureEnv} agent for both \textit{softlock} and \textit{hardlock} on all eligible joints. Both locking methods can limit the joint motion and show a similar pattern as the simulation results in Fig.~\ref{fig:virtual_joint}. Due to the manipulation of target joint position, \textit{softlock} tends to have much lower torque, and sometimes, the joint will overshot the locking range, making it less ideal and dangerous. But it is still sufficient for real-world validation on joints where \textit{hardlock} cannot be applied. For \textit{hardlock} on calf joint, the motion is limited to a range of around 0.15 rad and the large torque pattern similar to the virtual deployment is observed, demonstrating that the designed locking mechanism is efficient for real-world testing.

We further measure the survival time in the real world with a maximum lifetime of 20 seconds in Table~\ref{table:physical_deployement}. For the built-in controller, due to the limitation of A1's high-level API, we only apply \textit{hardlock}. The fault-tolerant \texttt{FailureEnv} agent can handle all the test seniors while the vanilla \texttt{BaseEnv} agent struggles on thigh and calf joint, which is in line with the observations of the virtual deployment in Sec.~\ref{section:virtual_failure}, and the robot stalls or falls directly to the ground. We further lock multiple joints for \texttt{FailureEnv} policy and the quadruped can still safely move forward even this situation is never seen during the training.

\begin{table}[t]
\caption{Survival time in physical tests under different joint locking}
\vspace{-10pt}
\label{table:physical_deployement}
\begin{center}
\begin{tabular}{c|ccc|c}
\Xhline{1.5pt}
\multirow{2}{*}{Agent} & \multicolumn{3}{c|}{Softlock} & \multirow{2}{*}{Hardlock} \\ \cline{2-4}
                       & Hip     & Thigh     & Calf    &                           \\ \Xhline{1.5pt}
FailureEnv             & 100\%       & 100\%         & 100\%       & 100\%                        \\
BaseEnv                & 100\%       & 20\%         & 5\%       & 35\%                         \\
Built-in               & -       & -         & -       & 0\%                        \\ \Xhline{1.5pt}
\end{tabular}
\vspace{-10pt}
\end{center}
\end{table} 

\section{Conclusion and Future Work}
In this paper, we propose a novel methodology to train and test hardware fault-tolerant RL-based controllers for quadruped locomotion, both in the simulation and real world. We design a novel simulation strategy for joint locking failures and a joint training pipeline to efficiently train a fault-tolerant quadruped locomotion controller with the teacher-student framework. We use commonly equipped low-level sensors available on quadrupedal robots as observations to generate robust actions. We demonstrate that even with one joint locked, our controller can still drive the quadruped without losing too much heading or speed. 


Quadrupedal robot failure is a complex topic and depends on the robot specifications. Delicate simulation and training are needed to be truly robust against all situations. Due to safety concerns, currently we only conduct experiments with forward movement, which can be further extended with additional user command input for controllable locomotion deployment. To further improve the fault tolerance of the locomotion controller, we will develop a unified and transferable solution for a variety of quadrupedal robotic platforms with different morphology, dynamics, and sensor sets. We can also pair additional high-level sensors such as depth cameras for safe and reliable task-related deployment. 


\bibliographystyle{IEEEtran}
\bibliography{bibtex/bib/IEEEabrv.bib,bibtex/bib/references.bib}{}

\end{document}